\documentclass[11pt]{article}
\usepackage{coling2020}

\usepackage{times}
\usepackage{url}
\usepackage{latexsym}
\usepackage{multirow}
\usepackage{multicol}
\usepackage{graphicx}
\usepackage{times}
\usepackage{latexsym}
\usepackage{amsfonts,amssymb}
\usepackage{booktabs}
\usepackage{amsmath}
\usepackage{multirow}
\usepackage{bm}
\usepackage{tcolorbox}
\usepackage{framed}
\usepackage{color}
\usepackage{colortbl}
\DeclareMathOperator*{\argmax}{argmax}

\DeclareMathOperator*{\concat}{concat}

\DeclareMathOperator*{\softmax}{softmax}

\DeclareMathOperator*{\avgpooling}{avg-pooling}
\usepackage{url}
\usepackage[dash,dot]{dashundergaps}
\usepackage[position=bottom]{subfig}

\usepackage{setspace}
\usepackage{array}

\usepackage{caption}

\definecolor{aqua}{rgb}{0.0, 1.0, 1.0}
\definecolor{babyblue}{rgb}{0.54, 0.81, 0.94}
\definecolor{ballblue}{rgb}{0.13, 0.67, 0.8}
\definecolor{bluegray}{rgb}{0.4, 0.6, 0.8}

\usepackage{hyperref}

\colingfinalcopy 

\title{Making the Best Use of Review Summary for Sentiment Analysis}

\author{Sen Yang$^{\heartsuit\triangle }$\hspace{0.5mm}\thanks{\hspace{1.5mm}Equal contribution.}\hspace{1.5mm}, 
Leyang Cui$^{\heartsuit\triangle \Diamond}$\hspace{0.5mm}$^*$, 
{\bf 
Jun Xie$^{\mathsection}$, 
Yue Zhang$^{\heartsuit\triangle}$\thanks{\hspace{1.5mm}Corresponding author}}\\
  $^\heartsuit$School of Engineering, Westlake University \\
  $^\triangle$Institute of Advanced Technology, Westlake Institute for Advanced Study\\
  $\Diamond$ Zhejiang University \hspace{6mm}
  $\mathsection$ Tencent SPPD\\
  \href{mailto:senyang.stu@gmail.com}{\tt senyang.stu@gmail.com} 
  \hspace{4mm} 
  \href{mailto:cuileyang@westlake.edu.cn}{\tt cuileyang@westlake.edu.cn}  \\
  \href{mailto:stiffxie@tencent.com}{\tt stiffxie@tencent.com} 
  \hspace{4mm} 
  \href{mailto:yue.zhang@wias.org.cn}{\tt yue.zhang@wias.org.cn} \\
  }

\date{}

\begin{document}
\maketitle
\begin{abstract}
Sentiment analysis provides a useful overview of customer review contents. 
Many review websites allow a user to enter a summary in addition to a full review. 
Intuitively, summary information may give additional benefit for review sentiment analysis. 
In this paper, we conduct a study to exploit methods for better use of summary information. 
We start by finding out that the sentimental signal distribution of a review and that of its corresponding summary are in fact complementary to each other. We thus explore various architectures to better guide the interactions between the two and propose a hierarchically-refined review-centric attention model. 
Empirical results show that our review-centric model can make better use of user-written summaries for review sentiment analysis, and is also more effective compared to existing methods when the user summary is replaced with summary generated by an automatic summarization system.
\end{abstract}

\blfootnote{This work is licensed under a Creative Commons Attribution 4.0 International Licence. Licence details: \url{http://creativecommons.org/licenses/by/4.0/}. }

\section{Introduction}
\label{sec:intro}
\begin{figure}[!t]
    \begin{center}
    \subfloat[A review for {\it Avalon} (a card game) from Amazon]{
    \includegraphics[width=0.42\textwidth]{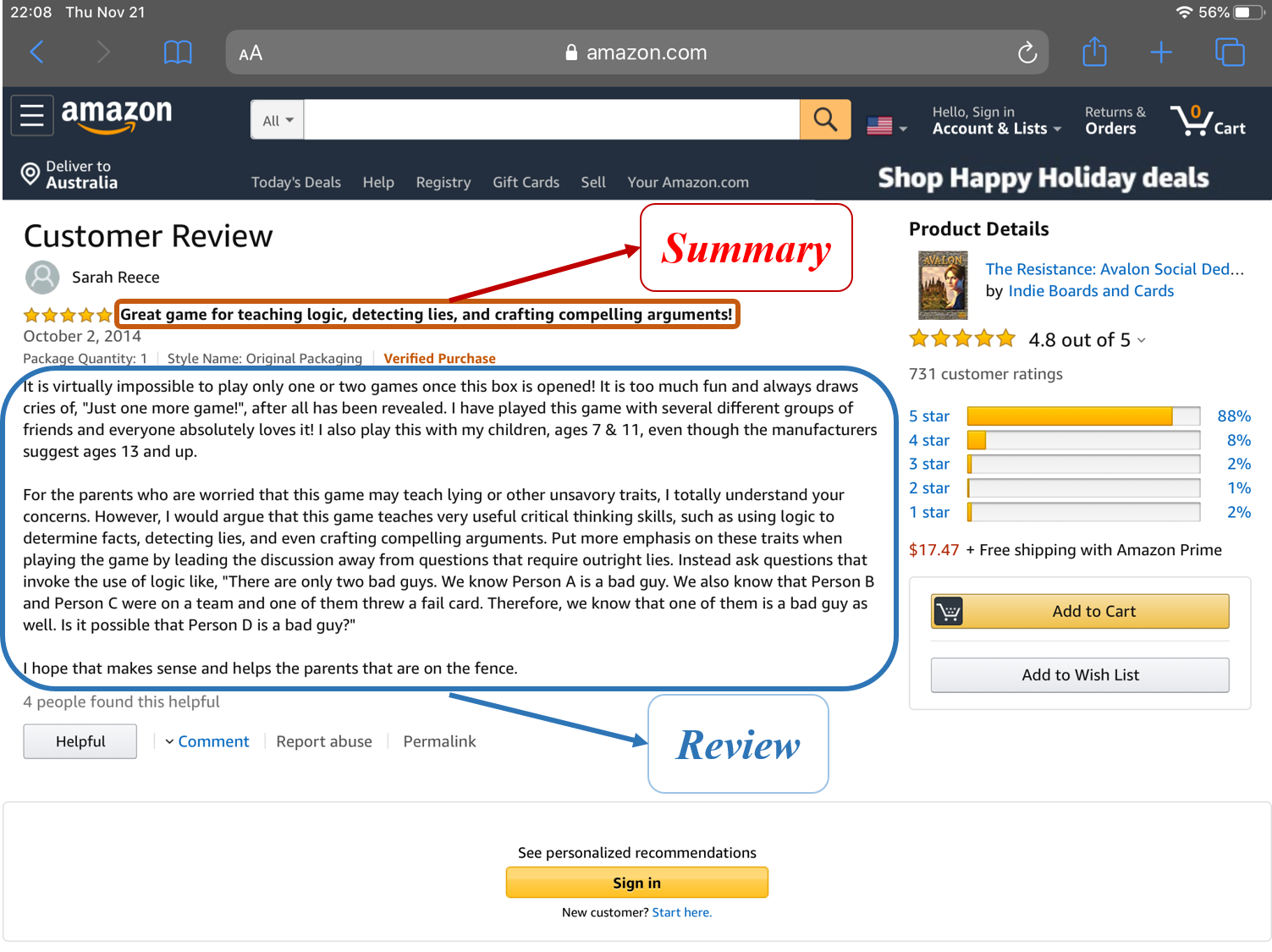}
    \label{fig:amazon_example}
    }\hspace{0.8cm}
    \subfloat[A review for the movie {\it Titanic} from IMDb]{
    \includegraphics[width=0.42\textwidth]{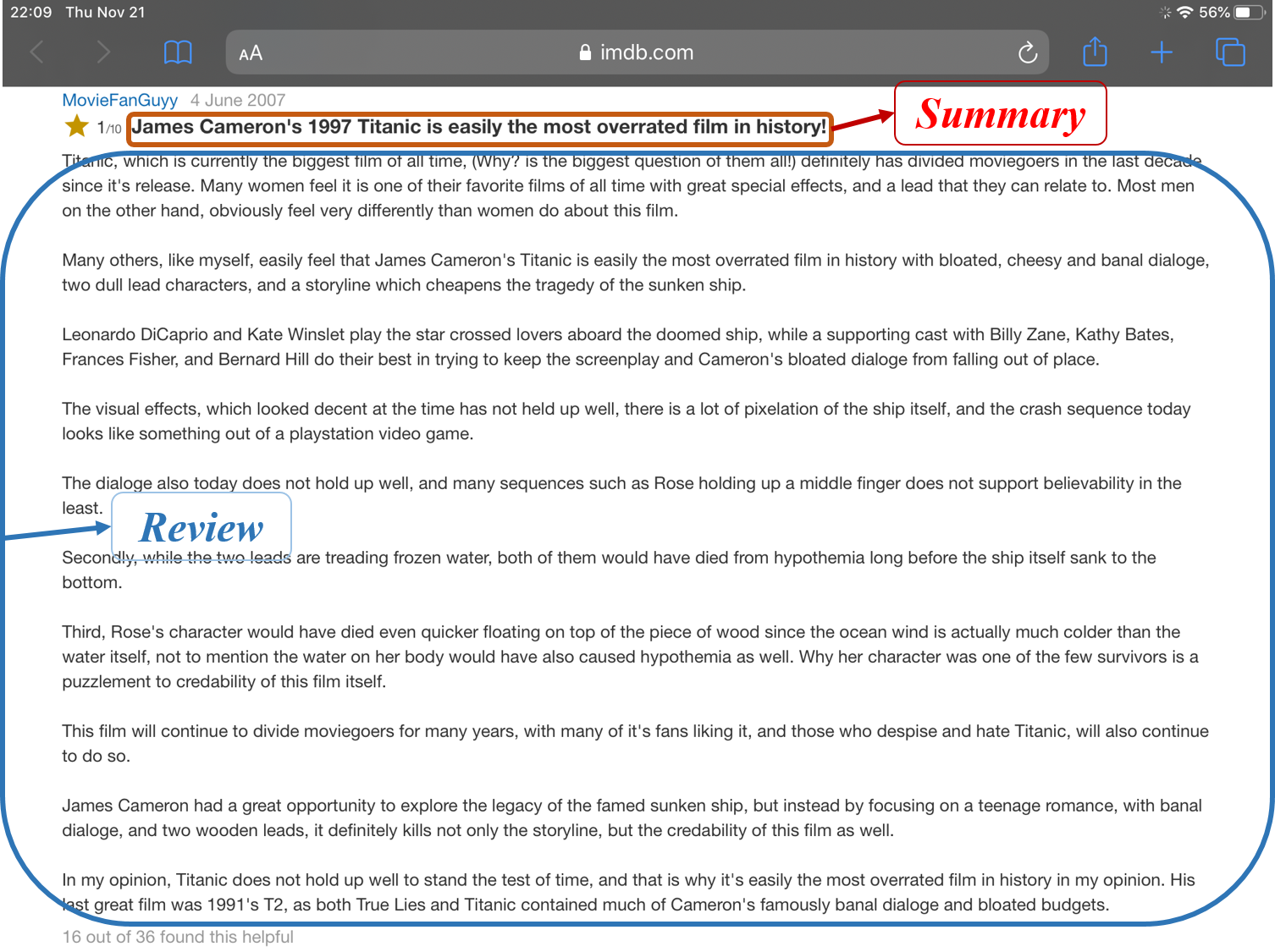}
    \label{fig:imdb_example}
    }
    \caption{Screenshots from two review websites, each containing a brief summary along with a review.}
    \label{fig:example}
    \end{center}
\end{figure}

Sentiment analysis~\cite{DBLP:conf/emnlp/PangLV02,kim-2004-determining,Liu:sentiment,DBLP:conf/emnlp/SocherPWCMNP13} is a fundamental task in natural language processing, which predicts the subjectivity and polarity of a given text. 
In practice, automatically extracting sentiment from user reviews has wide applications such as E-commerce and movie reviews~\cite{Manek2015AspectTE,DBLP:conf/ijcai/GuanCZZTC16,Kumari16}. 
In many review websites such as Amazon and IMDb, the user is allowed to give a summary in addition to the review, where summaries can contain more general information about the review. 
Figure~\ref{fig:example} gives a few such examples. 
It is thus an interesting research question on how to make use of both review and summary information for better sentiment classification under such a scenario. 

\begin{figure*}[!t]
\centering
\subfloat[Multi-task 
\label{fig:structure_1}]{\includegraphics[width=0.23\textwidth]{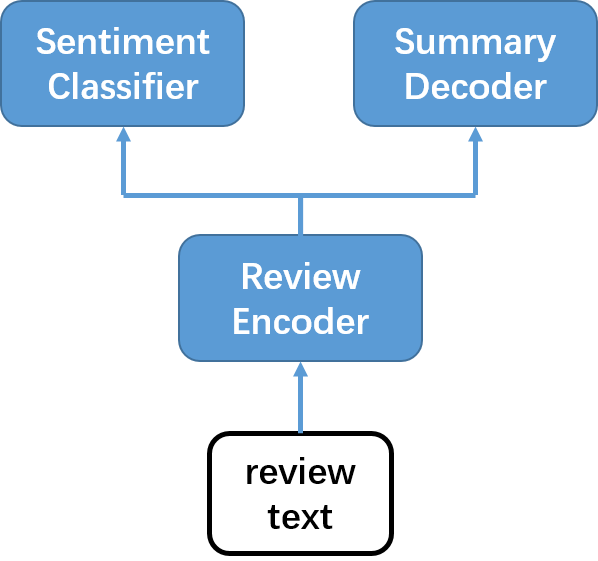}}
\hspace{1.0cm}
\subfloat[Separate encoder
\label{fig:structure_2}]{\includegraphics[width=0.23\textwidth]{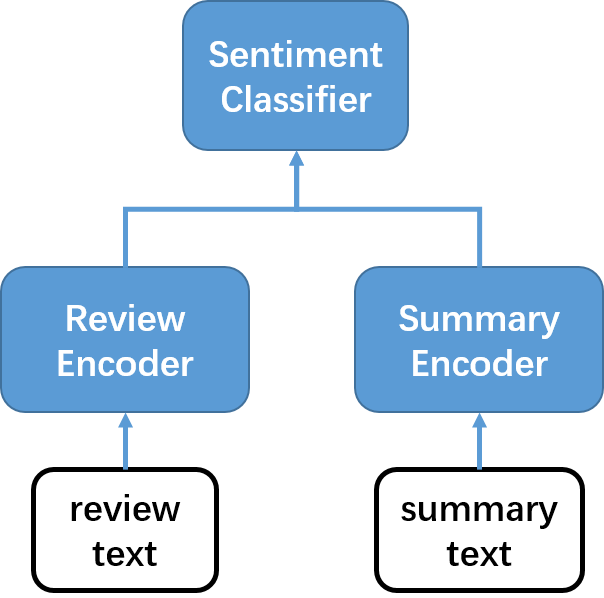}}
\hspace{1.0cm}
\subfloat[Joint encoder
\label{fig:structure_3}]{\includegraphics[width=0.23\textwidth]{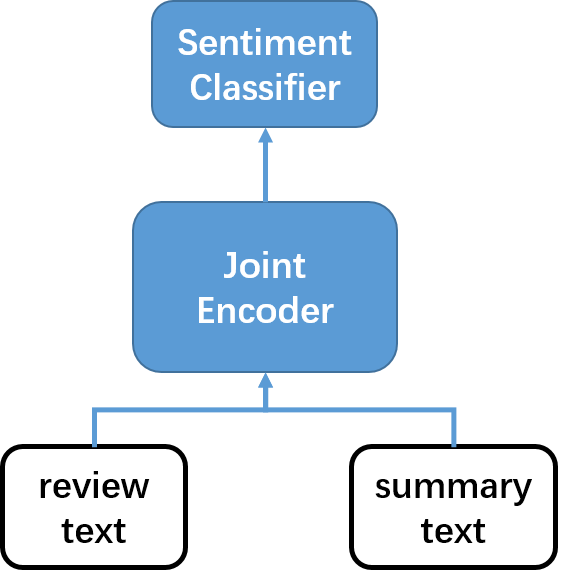}}
\caption{Three model structures for incorporating summary into sentiment classification. }
\label{fig:three_structure}
\end{figure*}

As shown in Figure~\ref{fig:example}, user-written summaries can be a brief version of reviews that is highly indicative of the user sentiment. Thus summaries can be used as additional training signals for sentiment classification. 
To this end, recent work~\cite{Ma2018bag,pmlr-v95-wang18b} exploits multi-task learning. The model structure can be illustrated by Figure~\ref{fig:structure_1}. In particular, given a review input, a model is trained to simultaneously predict the sentiment and the summary. As a result, both summary and review features are integrated into the review encoder through back-propagation training. 

\begin{table}[t]
\centering
        \begin{tcolorbox}[colframe=bluegray, colback=white, left = 1mm, right = 1mm, top = 0.5mm, bottom = 0.5mm, boxsep = 0mm,
      toptitle = 0mm, bottomtitle = 0mm, title = {\scriptsize Rating: 5 stars}, middle=0.5pt]
        \begin{spacing}{0.5}
        {\scriptsize\bf Review:} \textit{\scriptsize I can color right along with my grandchildren, without feeling intellectually compromised at the project. This book is \underline{\bf \textit{so amazing}} that I have used some of the designs for stained glass windows. I \underline{\bf \textit{highly recommend}} this for anyone who does not want to grow out of a favorite past time.}
        \end{spacing}
        \tcblower
        {\scriptsize\bf Summary:} \textit{\scriptsize Now I don't have to grow up}
        \end{tcolorbox}
          
        \begin{tcolorbox}[colframe=bluegray, colback=white, left = 1mm, right = 1mm, top = 0.5mm, bottom = 0.5mm, boxsep = 0mm,
      toptitle = 0mm, bottomtitle = 0mm, title = {\scriptsize Rating: 5 stars}, middle=0.5pt]
        \begin{spacing}{0.5}
        {\scriptsize\bf Review:} \textit{\scriptsize My son, 9, had outgrown his old helmet, so I bought this one. Less than three weeks later, he put it to the test. He is a daredevil who loves speed. Riding down a hill, that he isn’t supposed to ride on, he lost control at ~30 mph and landed on the side of his head, on the asphalt. He was knocked out momentarily, but passed the concussion screening at the ER. He is fine, other than some road rash. I hate to think how he might have fared with his old helmet, or no helmet. The ER doctor said that the spot he hit was about the worst place to hit for head injuries. Don’t skimp on safety equipment, ever, especially for kids. I am ordering this exact same helmet as a replacement.}
        \end{spacing}
        \tcblower
        {\scriptsize\bf Summary:} \textit{\scriptsize \underline{\bf \textit{Great buy!}} Saved my son, thank you}
        \end{tcolorbox}
    \caption{Two examples of online reviews with summaries and ratings. Explicit sentiment phrases are in bold and underlined.}
    \label{tab:example_2}

\end{table}

While the above methods are highly effective, we find that the correlation between reviews and summaries can be subtle. As shown in Table~\ref{tab:example_2}, sometimes a summary does not directly convey sentiment as contained in the review itself. In other cases, the summary contains explicit sentiment, but the review does not. Empirically, we find in our experiments that the sentiment polarities as predicted from the reviews are consistent with those predicted from the summaries for only $73.9\%$ of the test instances. 
Existing joint training methods take only the review as input at test time, and thus can be limited in its use of summary information.
These facts suggest that it can be necessary to model deeper interaction between reviews and summaries for better sentiment classification.

We conduct our investigation by treating both the review and the summary as inputs. 
In particular, we first compare the performance of sentiment classification using review only and using summary only, finding that the two sources of information are in fact complementary to each other. 
Second, as shown in Figure~\ref{fig:structure_2}, we investigate a simple method to integrate review and summary information by concatenating separately-learned representations. 
This method turns out to outperform models using review or summary inputs only.
One limitation of this method, however, is that it does not capture the interaction between the review and summary information as thoroughly as the method shown in Figure~\ref{fig:structure_1}, in which the representation of a review contains summary knowledge also.

To address this issue, we further investigate a joint encoder structure between the review and the summary, which is demonstrated in Figure~\ref{fig:structure_3}. 
To this end, an intuitive method is co-attention~\cite{iclr17-coattention}, which iteratively updates the representation of review and summary by consulting each other, as shown in Figure~\ref{fig:coattention-structure}.  
However, we find empirically that the review itself is relatively more indicative of the user sentiment compared to the summary. 
Given this observation, we further build a review-centric joint encoder. 
Different from the co-attention encoder, the review-centric model iteratively updates a review representation given a summary representation, but not vice versa.

We evaluate our proposed models on the SNAP (Stanford Network Analysis Project) Amazon review datasets \cite{He:2016:UDM:2872427.2883037}, which contain reviews and ratings together with \mbox{user-written} summaries if they exist. In scenarios where there is no user-written summary for a review, we use a pointer-generator network summarization model \cite{see2017pointer} to generate abstractive summaries. Empirical results show that our review-centric model outperforms a range of baselines, including multi-task, separate encoder and joint encoder methods.
In addition, our review-centric model achieves new state-of-the-art results, giving 2.1\% (with system-generated summary) and 4.8\% (with gold summary) absolute improvements compared to the previous best method on the SNAP benchmark. 
To our knowledge, we are the first to investigate the correlation between reviews and their summaries for expressing sentiment, and the first to empirically investigate different models making use of both reviews and summaries for better sentiment analysis. 
We release our code at \url{https://github.com/RingoS/sentiment-review-summary}.

\section{Related Work}
\label{related}
Our work is partly related to previous work building well-designed matching models to capture the relationship between two texts. 
In reading comprehension, a matching model is required to capture the similarity among a given passage, a question and a candidate answer. 
\newcite{chen-etal-2016-thorough} adopted two GRUs to encode the passage and question, respectively, and a bilinear function to compute the similarity on each passage token. 
\newcite{iclr17-coattention} make use of co-attention, which shares one single attention matrix between the passage and the question, calculating both passage-to-question and question-to-passage attention scores.
For retrieval-based dialogue systems, models are required to calculate the matching score between a candidate response and a conversation context. 
In particular, {\it Sequential Matching Network}~\cite{smn} captures matching information by constructing word-to-word and a sequence-to-sequence similarity matrices. 
{\it Deep Attention Matching Network} ~\cite{zhou-etal-2018-multi} adopts self-attention and cross-attention modules to harvest intra-sentence relationship and inter-sentence relationship, respectively.
To capture potential long-term label dependency in sequence labeling, \newcite{cui-zhang-2019} use attention over label embeddings to refine the marginal label probabilities by calculating the similarity between a word sequence and a set of label embeddings.
Compared with these methods, which model {\it matching} between two pieces of text, 
our work is different in that we consider how to effectively make use of the {\it complementary} property between a review and a corresponding summary for better review sentiment analysis.

Our work is related to previous work on sentiment analysis~\cite{DBLP:conf/emnlp/PangLV02,kim-2004-determining,Liu:sentiment}, taking a whole review as input~\cite{Kim14cnn,charCNN,YangYDHSH16hierarchical,johnson-zhang-2017-deep} rather than specific aspects~\cite{chen-etal-2017-recurrent-attention,li2019unified}. Different from previous work, we additionally consider user-generated or automatically-generated summaries as input. 
Our work is related to existing work on joint summarization and sentiment classification. 
\newcite{Ma2018bag} propose a multi-view attention model for joint summarization and sentiment classification. 
\newcite{pmlr-v95-wang18b} improve the model of \newcite{Ma2018bag} by using additional attention on the generated text. 
Different from their work, we are not directly concerned about making better summaries. Instead, we make a broader discussion on how to make the best use of both review and summary for sentiment classification. 
Our work is also related to rationalizing sentiment predictions. \newcite{zhang-etal-2016-rationale} regard gold-standard rationales as additional input and used rationale-level attention for text classification. \newcite{bastings2019} propose an unsupervised latent model that selects a rationale and subsequently uses it for sentiment analysis.  Our work is similar in that we can visualize the most salient words in sentiment classification. Different from the existing methods, our rationalization is based on the interaction between a review and a summary, with the latter guiding the visualization.

\section{Task}

\subsection{Problem Formulation}
The input to our task is a pair $(X^w, X^s)$, where $X^w = x^w_1, x^w_2, ..., x^w_n$ is a review and $X^s = x^s_1, x^s_2,...,x^s_m$ is a corresponding summary, the task is to predict the sentiment label $y \in [1, 5]$, where $1$ denotes the most negative sentiment and $5$ denotes the most positive sentiment. $n$ and $m$ denote the size of the review and summary in the number of words, respectively. 

\subsection{Research Questions}
\label{sec:question}
We aim to answer the following research questions empirically:
\begin{itemize}
    \item[$\circ$] RQ \#1: What are the roles of and the correlation between a review and its summary for predicting the user rating;
    \item[$\circ$] RQ \#2: How to better leverage information from both the review and the summary for effective sentiment classification;
\end{itemize}

\begin{figure*}[!t]
\centering
\subfloat[Architecture of a co-attention joint encoder
\label{fig:coattention-structure}]{\includegraphics[width=0.48\textwidth]{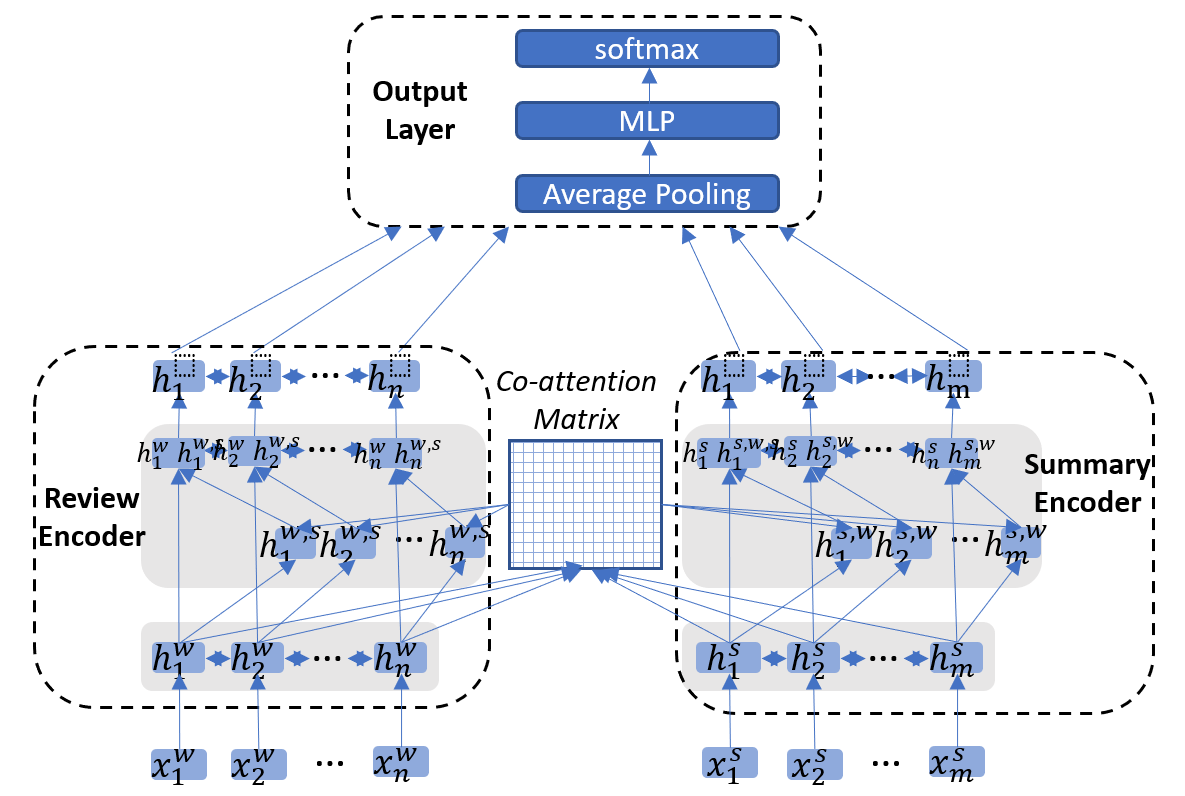}}
\subfloat[Architecture of a review-centric joint encoder
\label{fig:our_model_overview}]{\includegraphics[width=0.48\textwidth]{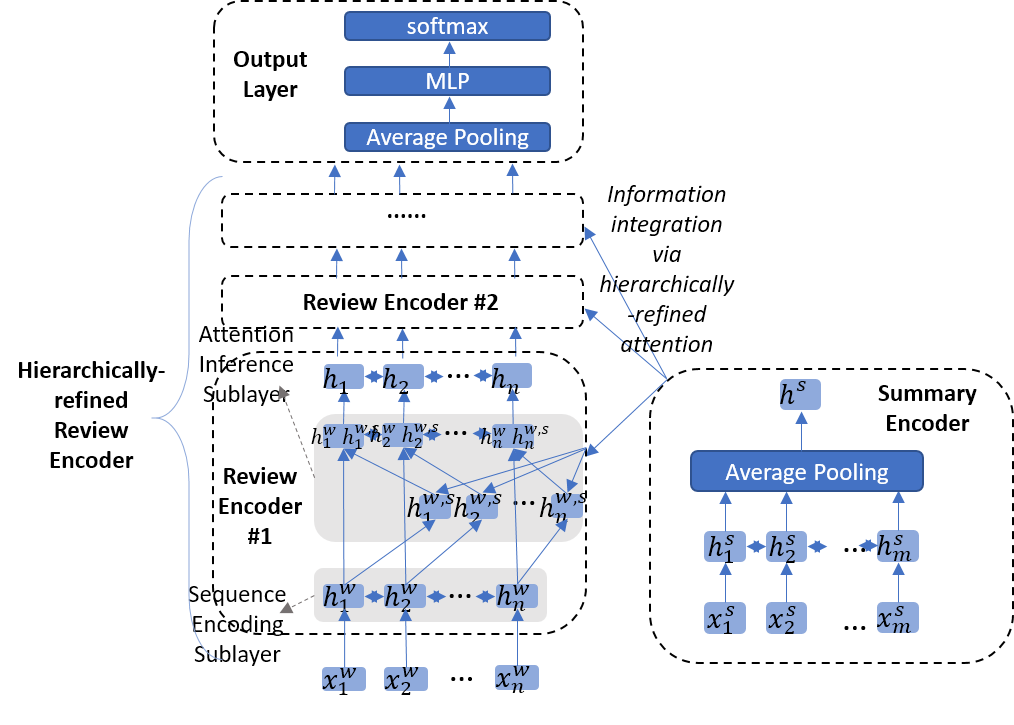}}
\caption{Architectures of co-attention model and our review-centric model. }
\label{fig:model_structure}
\end{figure*}

\section{Method}
\label{sec:method}
All the methods that we investigate are based on a BiLSTM~\cite{HochSchm97} structure. We first discuss the basic BiLSTM to encode text (Sec~\ref{sec:sequenceencoding}), and then discuss two types of structures that use BiLSTM for separate encoding(Sec~\ref{sec:separate-encoding}) and symmetric joint encoding (Sec~\ref{sec:symmetric}), respectively. Finally, we discuss review-centric joint encoding (Sec~\ref{sec:review-centric}) of the review and the summary.

\subsection{Sequence Encoding}
\label{sec:sequenceencoding}
We use BiLSTM as the sequence encoder for all experiments.
The input is a sequence of word representations $\mathbf{x} = \mathbf{x}_1, \mathbf{x}_2, ..., \mathbf{x}_m = \{emb(x_1), ..., emb(x_m)\}$, where $emb$ denotes a word embedding lookup table. Word representations are fed into a standard BiLSTM. We adopt a standard LSTM formulation, where a sequence of hidden states $\mathbf{h}_t$ are calculated from the sequence of $\mathbf{x}_t$~($t \in [1,...,m]$). 
 
A forward left-to-right LSTM layer and a backward right-to-left LSTM yield a sequence of forward hidden states $\{{\stackrel {\rightarrow}{\mathbf{h}_1}},...,{\stackrel {\rightarrow}{\mathbf{h}_m}}\}$ and a sequence of backward hidden states $\{{\stackrel {\leftarrow}{\mathbf{h}_1}},...,{\stackrel {\leftarrow}{\mathbf{h}_m}}\}$, respectively. The two hidden states are concatenated to form a final representation: 
\begin{equation}
\begin{split}
\mathbf{h}_i &= [\stackrel{\rightarrow}{\mathbf{h}_i};\stackrel{\leftarrow}{\mathbf{h}_i}]  \\
\mathbf{H} &= \{\mathbf{h}_1,..., \mathbf{h}_m\}
\end{split}
\end{equation}

This encoder structure serves as a basis for all the models. 
In particular, for the review-only and summary-only baselines, we use a single encoder as described above. 

\subsection{Baselines}
\label{sec:baseline}

\subsubsection{Separate Encoding}
\label{sec:separate-encoding}

Two BiLSTMs are adopted to separately encode reviews and summaries. 
Both of the produced hidden state matrices are then delivered to two settings:

1)\ \ average-pooling baseline: the hidden state matrices are concatenated and then average-pooled to form a final representation for later prediction.

2)\ \ self-attention baseline: the hidden state matrices are separately processed using self-attention mechanism. Subsequently the two matrices are concatenated and average-pooled to produce the final representation for later prediction. Our self-attention module follows the implementation of \newcite{lin2017structured}.

\subsubsection{Symmetric Joint Encoding}
\label{sec:symmetric}
On top of the sequence encoder, we separately adopt average pooling, self-attention~\cite{lin2017structured}, hard-attention~\cite{xu-icml2015,shankar-etal-2018-surprisingly} and co-attention~\cite{iclr17-coattention} mechanisms as our joint baselines. 
In particular, \mbox{co-attention} can capture the interactions between review and summary by calculating the bidirectional symmetric attention flows with a shared attention weight matrix. 


1)\ \ For joint encoder baselines using pooling and self-attention, only one BiLSTM is adopted, with concatenated review and summary texts as input. 

2)\ \ The hard-attention baseline is trained using an additional extractive summarization objective. 
We implement our baseline following \newcite{xu-icml2015} and \newcite{shankar-etal-2018-surprisingly}. 
In particular, words in the review text that overlap with the corresponding summary are extracted in their original order to formulate a summary.
The model calculates an additional loss between attention weights and extractive summary labels, so that the hard attention weights can be automatically produced during inference time. 

3)\ \ As for co-attention baselines, we use two BiLSTMs to separately encode review and summary. The two hidden state matrices then interact with each other. The formulations are written as :
\begin{equation}
\begin{split}
\mathbf{A}\ \ \ \ &= \mathbf{H}^w\mathbf{W}^w (\mathbf{H}^s\mathbf{W}^s)^{\intercal} \\
\mathbf{H}_{co\text{-}att}^{w} &= \mathbf{H}^w + \softmax(\frac{\mathbf{A}}{\sqrt{d}})\mathbf{H}^s \\
\mathbf{H}_{co\text{-}att}^{s} &= \mathbf{H}^s + \softmax(\frac{\mathbf{A}^{\intercal}}{\sqrt{d}})\mathbf{H}^w
\end{split}
\label{eqa:co-attention}
\end{equation}
where $\mathbf{H}^{w}$ and $\mathbf{H}^{s}$ are the hidden states of reviews and summaries, respectively. 
$\mathbf{A} \in \mathbb{R}^{n\times m}$ is the co-attention matrix. 
$n$ and $m$ are the lengths of the review text and the summary text, respectively. 
$d$ represents the hidden size of the BiLSTM. 
$\mathbf{H}_{co\text{-}att}^{w}$ and $\mathbf{H}_{co\text{-}att}^{s}$ are the \mbox{co-attention} representations of a review and its corresponding summary, respectively. They are then fed into subsequent layers for making predictions. 

\subsection{Review-centric Joint Encoding}
\label{sec:review-centric}
This joint encoder model changes the review encoder over the baseline, while keeping the summary encoder.
As shown in Figure~\ref{fig:our_model_overview}, the review encoder has a set of stacked layers, each consisting of a sequence encoding sublayer and an attention inference sublayer. The sequence encoding sublayers takes the same BiLSTM structure as the summary encoder, but with different model parameters. The attention inference sublayer integrates summary information into the review representation. By repeatedly consulting summary information, the review encoder obtains increasingly refined hidden states over layers.

\paragraph{Attention Inference Sublayer}
Formally, $X^s$ and $X^w$ are fed into sequence encoding layer, yielding $\mathbf{H}^w$ and $\mathbf{H}^s$, respectively. 
$\mathbf{h}^s$ is then obtained by average-pooling over $\mathbf{H}^s$.
We model the dependencies between the original review and the summary with multi-head dot-product attention. 
Each head produces an attention vector $\bm{\alpha} \in \mathbb{R}^{n}$, 
which consists of a set of similarity scores between the hidden state of each token of the review text and the summary representation. The hidden states are calculated by

\begin{equation}
\begin{split}
\bm{\alpha} \quad &= \softmax (\frac{\mathbf{H}^w \mathbf{W}_i^Q (\mathbf{h}^s \mathbf{W}_i^K)^{\intercal}}{\sqrt{d_h/k}}) \\
\mathbf{head}_i &= \mathbf{\hat{A}} ({\mathbf{\hat{H}}^s})^{\intercal} \mathbf{W}_i^V \\
\mathbf{H}^{w,s} &= \concat(\mathbf{head}_1,\dots,\mathbf{head}_k),  \\
\end{split}
\label{eqa:multi-head attention}
\end{equation}
where superscripts $w$ and $s$ represent review and summary, respectively. 
$\mathbf{\hat{A}}$ and ${\mathbf{\hat{H}}^s}$ are the unsqueezed matrices of $\bm{\alpha}$ and $\mathbf{h}^s$. 
$\mathbf{W}_i^Q \in \mathbb {R}^{d_{h} \times \frac{d_{h}}{k}}$, $\mathbf{W}_i^K \in \mathbb {R}^{d_{h} \times \frac{d_{h}}{k}}$ and $\mathbf{W}_i^V \in \mathbb {R}^{d_{h} \times \frac{d_{h}}{k}}$ are model parameters. 
$Q$, $K$ and $V$ represent \textit{Query}, \textit{Key} and \textit{Value}, respectively. 
$k$ is the number of parallel heads and $i \in [1,k]$ indicates which head is being processed. 

Following~\newcite{Transformer2017}, we adopt a residual connection around each attention inference layer: 
\begin{equation}
\mathbf{H} = {\rm{LayerNorm}}(\mathbf{H}^w + \mathbf{H}^{w,s})
\end{equation}

$\mathbf{H}$ is then fed to the subsequent sequence encoding layer as input, if any. 

According to the equations of standard LSTM and Equation \ref{eqa:multi-head attention}, tokens of the original review that are the most relevant to the summary are particularly focused on by consulting summary representation. 
The hidden states $\mathbf{H}^{w,s}$ are thus a representation matrix of the review text that encompasses key features of summary representation. 
Multi-head attention ensures that multi-faced semantic dependency features can be captured, which is beneficial for scenarios where multiple key points exist in one review.

\subsection{Output Layer}

Global average pooling is applied on $\mathbf{H}$, followed by a classifier layer:
\begin{equation}
\begin{split}
\mathbf{h}^{avg} &= \avgpooling(\mathbf{h}_1,...,\mathbf{h}_n)\\
\mathbf{p} \ \ &= \softmax(\mathbf{W} \mathbf{h}^{avg}+\mathbf{b}) \\
\hat{y} \ \ &=\argmax \mathbf{p}, 
\end{split}
\end{equation}
where $\hat{y}$ is the predicted sentiment label; $\mathbf{W}$ and $\mathbf{b}$ are parameters to be learned.

\paragraph{Training}
Given a dataset $D={\{(X^w_t,X^s_t,y_t)\}}|^{|T|}_{t=1}$, our models can be trained by minimizing
\begin{equation}
    L = - \sum_{t=1}^{|T|} \log(\mathbf{p}^{[y_t]})
\end{equation}
where $\mathbf{p}^{[y_t]}$ denotes the value of the label in $\mathbf{p}$ that corresponds to $y_t$.

\section{Experiments}
\label{sec:experiments}

\begin{table}[!t]
\begin{center}
\resizebox{0.48\textwidth}{!}{
\begin{tabular}{cccc}
\hline
\bf{Domain} &\bf{Size} & \bf{\#Review} & \bf{\#Summary} \\
\hline
Toys \& Games & 168k & 99.9 & 4.4 \\
Sports \& Outdoors & 296k & 87.2 &4.2 \\
Movies \& TV & 1,698k & 161.6 & 4.8 \\
\hline
\end{tabular}
}
\end{center}
\caption{\label{dataset} Data statistics. Size: number of samples, \#Review: the average length of reviews, \#Summary: the average length of summaries.}
\label{tab:dataset}
\end{table}

The SNAP Amazon Review Dataset\footnote{\url{http://snap.stanford.edu/data/web-Amazon.html}}~\cite{McAuley2013} consists of around 34 million Amazon reviews in different domains, such as books, games, sports and movies. Each review mainly consists of a product ID, a piece of user information, a plain text review, a user-written summary and an overall sentiment rating which ranges from 1 to 5. 
For fair comparison with previous work, we adopt the same domains and partition used by~\newcite{Ma2018bag} and \newcite{pmlr-v95-wang18b}, which includes three datasets ({Toys \& Games}, {Sports \& Outdoors} and {Movies \& TV}). 
The statistics of our adopted dataset are shown in Table \ref{tab:dataset}.
For each dataset, the first 1000 samples are taken as the development set, the next 1000 samples as the test set, and the rest as the training set.


\begin{table}[!t]
\begin{center}    
\begin{minipage}{0.46\textwidth}
\centering
\resizebox{\textwidth}{!}{
\begin{tabular}{ccccc}
\hline
\bf{Model} & \bf{T \& G} & \bf{S \& O} & \bf{M \& T} & {\bf Average}\\
\hline
\multicolumn{5}{c}{Multi-task} \\
\hline
HSSC &  71.9 & 73.2 & 68.9 & 71.3 \\
SAHSSC &  72.5 & -- & 69.2 & 70.9 \\
\hline
\multicolumn{5}{c}{Summary Only} \\
\hline
Pooling & 73.0 & 69.5 & 68.1 & 70.2 \\
Self-attention & 71.9 & 70.4 & 68.9 & 70.4 \\
\hline
\multicolumn{5}{c}{Review Only} \\
\hline
Pooling &  73.3 & 71.2 & 71.7 & 72.1 \\
Self-attention &  73.5 & 71.8 & 72.3 & 72.5 \\
\hline
\multicolumn{5}{c}{Separate Encoder} \\
\hline
Pooling & 74.4 & 73.9 & 73.8 & 74.0 \\
Self-attention & 75.8 & 73.1 & 73.7 & 74.2 \\
\hline
\multicolumn{5}{c}{Joint Encoder} \\
\hline
Hard-attention & 73.4 & 72.1 & 73.9 & 73.1 \\
Pooling & 75.4 & 73.4 & 73.2 & 74.0 \\
Self-attention & 75.7 & 74.3 & 74.1 & 74.7 \\
Co-attention & 76.1 & 74.2 & 74.3 & 74.9 \\
\bf{Review-centric Model} & \bf{76.6} & \bf{76.1} & \bf{75.9} & {\bf 76.2}\\
\hline
\end{tabular}
}
\caption{Results using gold summary as input.}
\label{tab:experiment-result:gold}
\end{minipage}\hfill
\begin{minipage}{0.46\textwidth}
\centering
\begin{minipage}{\textwidth}
\resizebox{\textwidth}{!}{
\begin{tabular}{ccccc}
\hline
\bf{Model} & \bf{T \& G} & \bf{S \& O} & \bf{M \& T} & {\bf Average}\\
\hline
\multicolumn{5}{c}{Separate Encoder} \\
\hline
Pooling & 71.8 & 72.2 & 72.5 & 72.2 \\
Self-attention & 73.1 & 72.5 & 72.6 & 72.7 \\
\hline
\multicolumn{5}{c}{Joint Encoder} \\
\hline
Pooling & 73.8 & 72.0 & 72.0 & 72.6 \\
Self-attention & 73.9 & 71.6 & 72.4 & 72.6 \\
Co-attention & 73.8 & 72.2 & 72.7 & 72.9 \\
\bf{Review-centric Model} & {\bf 74.8} & {\bf 72.6} & {\bf 72.8} & {\bf 73.4} \\
\hline
\end{tabular}
}
\caption{Results using system-generated summary as input.}
\label{tab:experiment-result:predicted}
\end{minipage}
\par
\begin{minipage}{\textwidth}
\resizebox{\textwidth}{!}{
\begin{tabular}{c|ccc|c} 
\hline
\bf{Model} & \bf{T \& G} & \bf{S \& O} & \bf{M \& T} & {\bf Average}\\
\hline
Co-attention (\textit{review}) & 75.1 & 74.8 & 74.7 & 74.9 \\
Co-attention (\textit{summary}) & 74.1 & 74.2 & 73.4 & 73.9 \\
Co-attention (\textit{concat}) & 76.1 & 74.2 & 74.3 & 74.9 \\
Summary-centric & 73.8 & 74.5 & 74.9 & 74.4 \\
Review-centric & {76.6} & {76.1} & {75.9} & { 76.2}\\
\hline
\end{tabular}
}
\caption{Result comparison among different interacting schemes.}
\label{tab:comparison-review-summary}
\end{minipage}
\end{minipage}
\end{center}
\end{table}


\subsection{Experimental Settings}
We use GloVe~\cite{pennington2014glove} \mbox{300-dimensional} embeddings as pretrained word vectors. The LSTM hidden size is set to 256. We use Adam~\cite{journals/corr/KingmaB14} to optimize all models, with an learning rate of $3e-4$, momentum parameters $\beta_1 = 0.9$, $\beta_2 = 0.999$, and $\epsilon = 1 \times 10^{-8}$. The dropout rate $\alpha$ and number of attention heads $M$ are separately set depending on the size of each dataset and using development experiments, which are $\alpha = 0.5$ and $M=1$ for {Toys \& Games}, $\alpha=0.2$ and $M=1$ for {Sports \& Outdoors} and $\alpha=0.0$ and $M=2$ for {Movies \& TV}. We adopt two layers for our review-centric model. 

In addition to conducting experiments with user-written summaries, we additionally perform experiments by replacing the user-written summary with a system-generated summary for two reasons. First, we want to know the extent to which our method can be generalized to settings of traditional sentiment classification, where the input consists of only one piece of text. This is the setting adopted by most previous research. Second, two of our baselines, namely HSSC~\cite{Ma2018bag} and SAHSSC~\cite{pmlr-v95-wang18b}, adopt this setting and use summary information via multi-task learning. 
For generating summaries, we separately adopt a pointer-generator network (PG-Net) with coverage mechanism~\cite{see2017pointer} trained on the training set. 

\subsection{Results}
Our main results are shown in Tables~\ref{tab:experiment-result:gold} and \ref{tab:experiment-result:predicted}. 
It can be seen from Table~\ref{tab:experiment-result:gold} that the baseline using review only outperforms that using summary only, which indicates that the review is more informative than the summary. 
In addition, the {\it Separate Encoder} models outperform both the {\it Summary Only} and the {\it Review Only} models, which indicates that additional summary input is beneficial to sentiment analysis. 
Finally, the {\it Joint Encoder} models generally outperform the {\it Separate Encoder} models, which suggests that modeling interactions between review and summary is superior to separate encoder structure. 
In particular, hard-attention receives more supervision information compared with soft-attention, by using supervision signals from {\it extractive} summaries. However, it underperforms the soft-attention model, which indicates that the most salient words for making sentiment classification may not strictly overlap with {\it extractive} summaries. 
Among soft-attention methods, co-attention achieves better performance compared to self-attention, which may result from the fact that co-attention allows mutual interactions between review and summary. 

In Table~\ref{tab:experiment-result:gold}, all architectures using system-generated summary as additional input outperform {\it Review Only} models, demonstrating that even imperfect summary can still give additional benefit for sentiment prediction. However, models using system-generated summary perform significantly worse than those using gold summary, verifying the importance of high quality summaries. 

In both tables, the review-centric model outperforms all the baseline models on all the three datasets. In particular, the review-centric model gives 1.3\% and 0.5\% improvements compared with the best baseline (co-attention) with both gold summary and system-generated summary, respectively. 
Our {\it Joint Encoder} models also outperform HSSC~\cite{Ma2018bag} and SAHSSC~\cite{pmlr-v95-wang18b}. In particular, these two multi-task models use summary information in training, thereby enhancing a review-only sentiment classifier. Their performance is competitive compared to the review only models. By further using user-written summaries directly, both {\it Separate Encoder} and {\it Joint Encoder} models outperform these models. It is worth noting that in Table~\ref{tab:experiment-result:predicted}, our methods still outperform the baselines with the same input settings, showing the effect of joint encoding. 


\subsection{Discussion}
In this section, we aim to answer the research questions raised in Section~\ref{sec:question}.

\subsubsection{RQ \#1}

\begin{figure}[!t]
\centering
\begin{minipage}[t]{.46\textwidth}
  \centering
  \includegraphics[width=\textwidth]{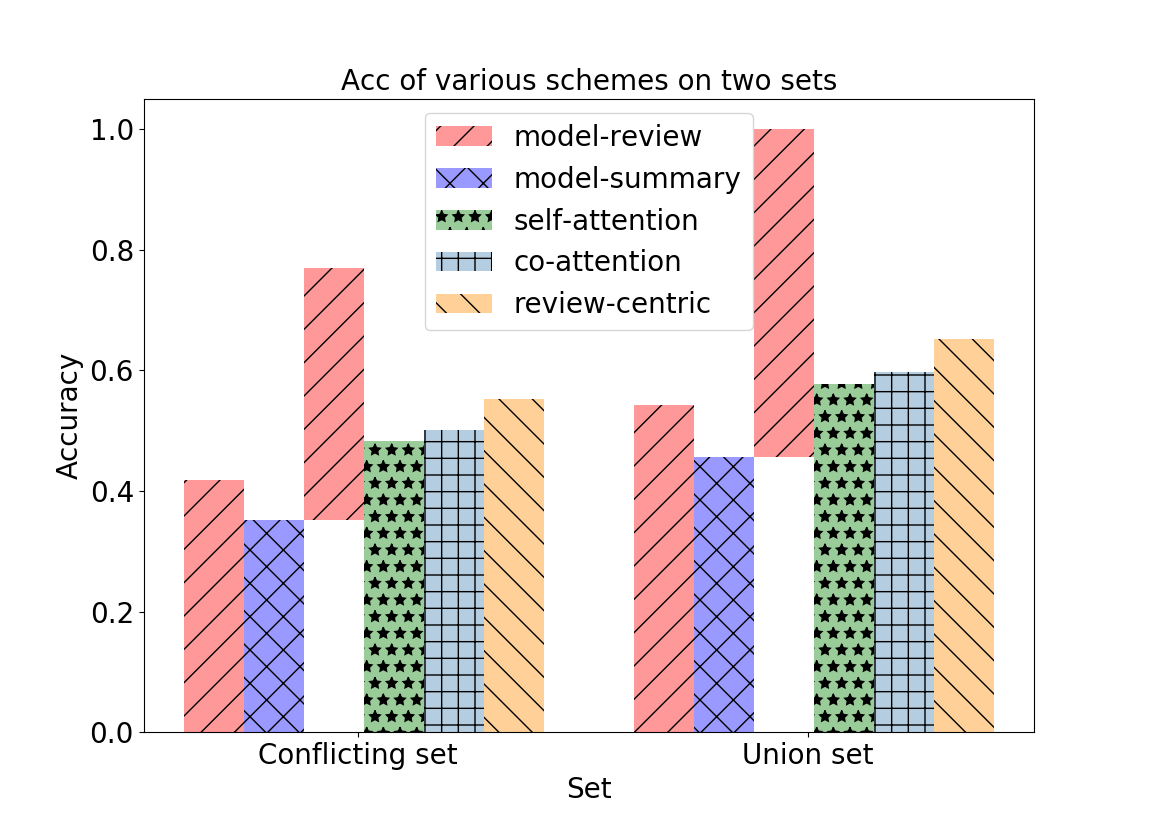}
  \captionof{figure}{Analysis on \textit{conflicting-set}. We stack the accuracy of \mbox{\textit{review-only}} and \mbox{\textit{summary-only}} to form the third bar, which is \textit{union-set}.}
  \label{fig:experiment-result:complementary}
\end{minipage}%
\hfill
\begin{minipage}[t]{.46\textwidth}
  \centering
  \includegraphics[width=\textwidth]{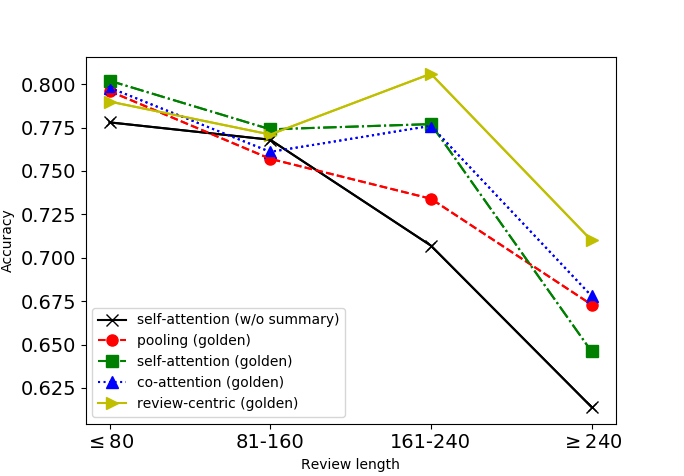}
  \captionof{figure}{Accuracy against the review length with gold summary input. }
  \label{fig:length_gold}
\end{minipage}
\end{figure}

We first explore the correlation between reviews and summaries with regard to carrying the user sentiment. 
In particular, we empirically compare the predictions of two simplest conditions, including {using review only} (abbreviated as \mbox{\textit{review-only}}) and {using gold summary only} (abbreviated as \mbox{\textit{summary-only}}), based on BiLSTM+pooling, on the \mbox{Toys \& Games} dataset. 
For the purpose of exploring correlation, we focus on a special part of the test set, named as {\it conflicting-set}, on which \mbox{\textit{review-only}} and \mbox{\textit{summary-only}} have conflicting predictions with each other. 
We assume that a review in {\it conflicting-set} contains a different sentiment rating from that of its corresponding summary.
{\it Conflicting-set} takes 26.1\% of the whole test set, which suggests that such {\it conflicting} samples are frequently seen in the dataset. 
Additionally, we define the complement of {\it conflicting set} as \mbox{\it non-conflicting-set}. 
We also define {\it union-set}, which is a subset of {\it conflicting-set} and is composed of the samples for which at least one of the two models (\textit{review-only} and \textit{summary-only}) has correct predictions. 
The experimental results are shown in Figure~\ref{fig:experiment-result:complementary}. 

\paragraph{Correlation}
As shown in Figure~\ref{fig:experiment-result:complementary}, the co-attention model gives a low accuracy of $50.1\%$ on {\it conflicting-set}, which is much lower compared to its performance on \mbox{\it non-conflicting-set} ($85.1\%$). This wide gap suggests that conventional models have difficulty handling conflicting situations. 
It can also be seen from Figure~\ref{fig:experiment-result:complementary} that both \textit{review-only} and \textit{summary-only} obtain poor performance on {\it conflicting-set} ($41.8\%$ and $35.2\%$, respectively). 
However, the sum accuracy of the two models, which forms the third bar on both sides of Figure~\ref{fig:experiment-result:complementary}, 
takes $41.8\% + 35.2\% = 77.0\%$ of {\it conflicting-set}, 
which suggests that review and summary information are highly complementary to each other under conflicting situations.

\subsubsection{RQ \#2}

\paragraph{Interacting Scheme}
As shown in Tables~\ref{tab:experiment-result:gold} and \ref{tab:experiment-result:predicted}, our review-centric model gives better results compared to the co-attention method. 
The only difference between these two methods is the interacting scheme between the review and the summary. 
We thus conduct experiments to further explore the influences of different interacting schemes. 
In particular, we train co-attention models using three types of top-layer representations, including using review representations only, using summary representations only and using the concatenation of both of the above. 
The experiment results are shown in Table~\ref{tab:comparison-review-summary}. 
It can be seen that co-attention ({\it review}) and co-attention ({\it concat}) give rather similar performance, which indicates the relative importance of review representation. 
Moreover, the review-centric model outperforms all \mbox{co-attention} methods, which indicates that our review-centric attention is better than symmetric attention (e.g. co-attention). 

In addition, we empirically compare our model with a model with the same structure but reverse inputs, named as the summary-centric model. As shown in Table~\ref{tab:comparison-review-summary}, our review-centric model outperforms the summary-centric model by a large margin, which suggests that focusing on the review side is better than focusing on the summary side for predicting sentiment ratings. 

\subsection{Analysis}

\begin{figure*}[!t]
\centering
\subfloat[Attention heatmap with system-generated summary \label{fig:case-predicted}]{\fbox{\includegraphics[width=0.92\textwidth]{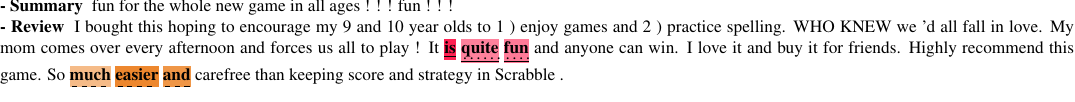}}}
\hspace{0.2cm}
\subfloat[Attention heatmap with gold summary \label{fig:case-gold}]{\fbox{\includegraphics[width=0.92\textwidth]{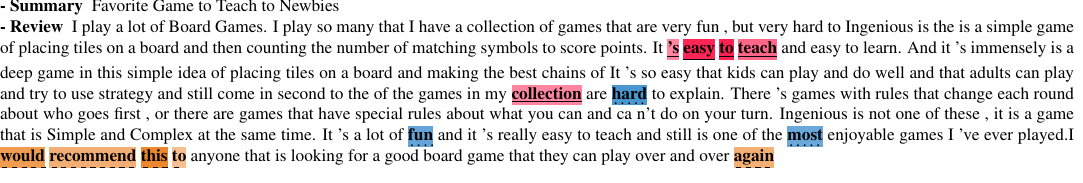}}}
\caption{Visualization of self-attention and hierarchically-refined attention, with system-generated summary (a) and gold summary (b). (1) \dotuline{BiLSTM+self-attention: dot line / blue color}; (2) \underline{the first layer of our review-centric model: straight line / pink color}; (3) \dashuline{the second layer of our review-centric model: dash line / yellow color}. Deeper color indicates higher attention weight. } \label{fig:case}
\end{figure*}

\paragraph{Intersection with Union-set}
We find that, on \textit{conflicting-set}, 
$92.1\%$ of the self-attention baseline's correct predictions, $91.0\%$ of the co-attention baseline's correct predictions and $91.0\%$ of the review-centric model's correct predictions come from \textit{union-set}. 
The line of high ratios suggests that explicit sentiment indication in at least one piece of text between the review and the summary is necessary for making a correct prediction on \textit{conflicting-set}.

In addition, our review-centric model slightly underperforms the co-attention model on \mbox{\it non-conflicting-set} ($84.2\%$ comparing with $85.1\%$). 
However, it still outperforms the co-attention model by $0.5\%$ on the whole Toys \& Games test set, which results from the fact that the the former outperforms the latter by a large margin of $5.1\%$ on \textit{conflicting-set}, and more specifically, $5.5\%$ on \textit{union-set}. 
The review-centric model's superior performance on \textit{union-set} verifies its strength on making better use of the complementary correlation between the review and the summary. 
It also suggests that the two models hold different inductive biases when encoding reviews and summaries. 

\paragraph{Review Length}
Figure~\ref{fig:length_gold} shows the accuracy of the average-pooling model, the self-attention model, the co-attention model and the review-centric model against review length. As the review length increases, the performance of all models decreases. 
BiLSTM+self-attention does not outperform BiLSTM+pooling on long text. 
Our review-centric method gives better results compared to all baseline models for long reviews, demonstrating that the review-centric model is effective for producing more abstract representations. 
The superior performance may result from the hierarchical review-centric attention mechanism, which maintains the most salient information while ignoring redundant information of the source review text. 
The review-centric model can thus be more robust when the review has noisy sentimental words or phrases, which are commonly seen in long reviews (e.g., the example in Figure~\ref{fig:case-gold}).

\paragraph{Case Study}
\label{sec:casestudy}

Our  models have a natural advantage of interpretability thanks to the use of the attention inference sublayer. 
We visualize the hierarchically-refined review-centric attention of two sample cases from the test set of Toys \& Games, and also self-attention distribution for fair comparison. 
To make the visualizations clear and to avoid confusion, we choose to visualize the most salient parts, by rescaling all attention weights into an interval of $[0, 100]$ and adopting $50$ as the threshold for attention visualization (only attention weights $\geq 50$ are visualized). 

Figure~\ref{fig:case-predicted} shows an example with system-generated summary that has 5 stars as the gold rating score. 
The summary text is ``{\it fun for the whole new game in all ages ! ! ! fun ! ! !}", which suggests that the game is 
1) interesting (from word ``{\it fun}") 
and 
2) not difficult to learn (from phrase ``{\it all ages}"). 
It can be seen that both the self-attention model and the first layer of our review-centric model attend to the strongly positive phrase ``{\it quite fun}", which is relevant to the word ``{\it fun}" in the summary. 
In comparison, the second layer attends to the phrase ``{\it much easier}", which is relevant to the phrase ``{\it in all ages}" in the summary. 
This verifies our review-centric model's effectiveness of leveraging abstractive summary information.

Figure~\ref{fig:case-gold} illustrates a 5-star-rating example with a gold summary. 
The summary text is ``{\it Favorite Game to Teach to Newbies}". 
As shown in the heatmap, self-attention attends only to general sentimental words such as ``{\it hard}", ``{\it fun}", ``{\it immensely}" and ``{\it most}", which deviates from the main idea of the document text. 
In comparison, the first layer of our review-centric model attends to phrases like ``{\it easy to teach}", which is a perfect match of the phrase ``{\it teach to newbies}" in the summary. 
This shows that the shallow attention inference sublayer can learn direct similarity matching information under the supervision of summarization. 
In addition, the second layer of our review-centric model attends to phrases including ``{\it would recommend this to anyone}", which links to ``{\it easy to teach}" and ``{\it Teach to Newbies}", showing that the deep attention inference sublayer of our model can learn underlying connections between the review and the summary.

\section{Conclusion}
We empirically analyzed the correlation between reviews and summaries for customer review sentiment analysis, found that they are complementary to each other for carrying user sentiment. 
We investigated a range of joint encoder models for better modeling the interactions between reviews and summaries and proposed a novel review-centric method, which hold different inductive bias to capture the complementary correlation. 
Empirical results verified the effectiveness of joint encoding for review and summary among strong baselines and existing work,  
showing that a review-centric model outperforms a symmetric co-attention model. 

\section*{Acknowledgments}
We thank all anonymous reviewers for their constructive comments. We also would like to acknowledge funding support from the Westlake University and Bright Dream Robotics Joint Institute for Intelligent Robotics and a research grant from Tencent.

\bibliographystyle{coling}
\bibliography{coling2020}


\end{document}